\pdfoutput=1

\documentclass[11pt]{article}

\usepackage[]{acl}

\usepackage{times}
\usepackage{latexsym}

\usepackage[T1]{fontenc}

\usepackage[utf8]{inputenc}

\usepackage{microtype}

\usepackage{inconsolata}
\usepackage{lipsum}
\usepackage{tcolorbox}
\usepackage{xcolor}
\usepackage{multicol}
\usepackage{enumitem}
\usepackage{catchfile}
\usepackage{booktabs}

\usepackage{changepage, threeparttable}
\usepackage{booktabs, multirow}
\usepackage{soul}
\usepackage{makecell}
\usepackage{enumitem}
\setlist{noitemsep}

\usepackage{array}
\usepackage{fullpage}
\usepackage{float}

\title{Publicly Shareable Clinical Large Language Model\\Built on Synthetic Clinical Notes}

\author{Sunjun Kweon$^{1}$$^{*}$, Junu Kim$^{1}$\thanks{\hspace{0.2cm} Equal contribution}, Jiyoun Kim$^{1}$, Sujeong Im$^{1}$, Eunbyeol Cho$^{1}$\\
    \textbf{Seongsu Bae$^{1}$, Jungwoo Oh$^{1}$, Gyubok Lee$^{1}$, Jong Hak Moon$^{1}$, Seng Chan You$^{2}$}\\
    \textbf{Seungjin Baek$^{2}$, Chang Hoon Han$^{2}$, Yoon Bin Jung$^{2}$, Yohan Jo$^{3}$, Edward Choi$^{1}$}\\
  KAIST$^{1}$ Yonsei University College of Medicine$^{2}$ Seoul National University$^{3}$ \\
  \texttt{\{sean0042, kjune0322, jiyoun.kim, sujeongim, eunbyeol.cho\}@kaist.ac.kr}\\
  \texttt{\{seongsu, gyubok.lee, jhak.moon, edwardchoi\}@kaist.ac.kr}\\
  \texttt{\{chandryou, bjh7790, paul9567, ybjung\}@yuhs.ac}\hspace{0.5cm}\texttt{\{yohan.jo\}@snu.ac.kr} \\ 
  }

\begin{document}
\maketitle
\begin{abstract}
The development of large language models tailored for handling patients’ clinical notes is often hindered by the limited accessibility and usability of these notes due to strict privacy regulations.
To address these challenges, we first create synthetic large-scale clinical notes using publicly available case reports extracted from biomedical literature.
We then use these synthetic notes to train our specialized clinical large language model, \textbf{Asclepius}.
While Asclepius is trained on \textit{synthetic data}, we assess its potential performance in real-world applications by evaluating it using \textit{real clinical notes}.
We benchmark Asclepius against several other large language models, including GPT-3.5-turbo and other open-source alternatives. 
To further validate our approach using synthetic notes, we also compare Asclepius with its variants trained on real clinical notes. 
Our findings convincingly demonstrate that synthetic clinical notes can serve as viable substitutes for real ones when constructing high-performing clinical language models. 
This conclusion is supported by detailed evaluations conducted by both GPT-4 and medical professionals. 
All resources—including weights, codes, and data—used in the development of Asclepius will be made publicly accessible for future research\footnote{\url{https://github.com/starmpcc/Asclepius}}.
\end{abstract}

\section{Introduction}

\begin{figure}
    \centering
    \includegraphics[width=0.5\textwidth, trim={150 200 130 150}, clip]{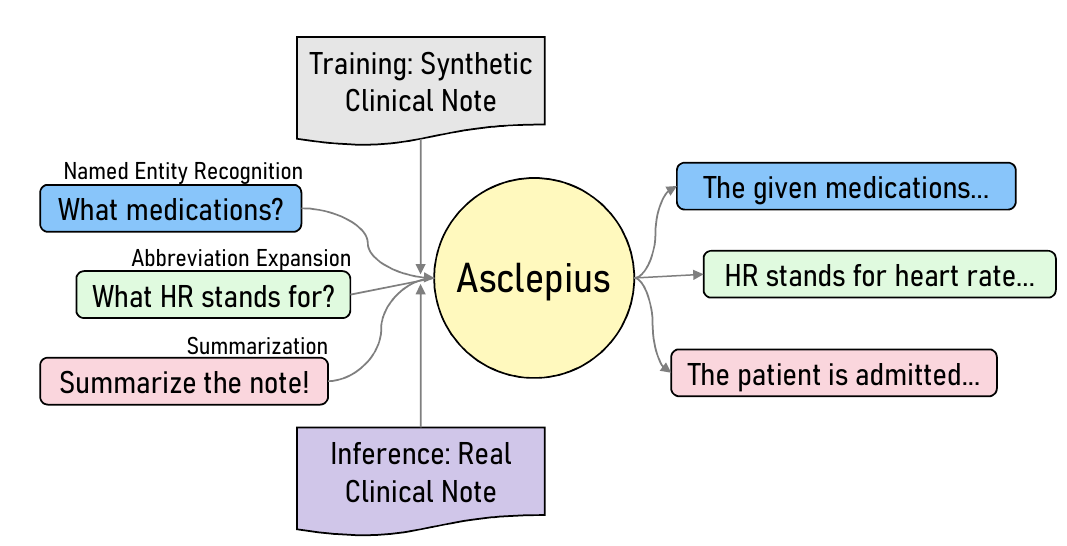}
    \caption{The large clinical language model, \textbf{Asclepius}, trained solely on synthetic clinical notes, can effectively handle various clinical NLP tasks on real notes in a zero-shot setting.}
    \label{fig:task}
    \vspace{-1em}
\end{figure}

Clinical notes serve as an extensive repository of information specific to individual patients. Applying Natural Language Processing (NLP) techniques to these notes can significantly enhance the decision-making processes of medical professionals \citep{demner2009can, lederman2022tasks, wu2022survey}.
Recent advances in large language models (LLMs) such as OpenAI's GPT series \citep{brown2020language,ouyang2022training,openai2023gpt4} have shown promising results in analyzing these clinical notes \citep{agrawal2022large, hu2023zero, liu2023deid, tang2023evaluating}.
However, when health organizations try to utilize these API-based external LLMs, they encounter two major challenges.

The first challenge is privacy and security.
Hospitals must transmit sensitive patient information beyond their internal systems when using these API-based external LLMs.
This could potentially infringe on privacy regulations.
Even when the external model adheres to regulations such as Health Insurance Portability and Accountability Act (HIPAA), hospitals should still undertake careful measures such as de-identifying clinical notes and setting up secure transmission protocols to avoid privacy breaches.
This complicates the usage of external models.
The second challenge relates to the autonomy that a health organization would need to exercise over its LLMs.
Given each organization's unique environment and characteristics, they may prefer a model specifically tailored to their needs.
In light of these challenges, there is an increasing demand for a clinical LLM that can operate securely in an offline environment while still offering the effectiveness of powerful online LLMs such as GPT series.

To develop a specialized LLM capable of handling clinical notes, a specific training dataset is required. 
This dataset would consist of instruction-answer pairs drawn from real clinical notes.
Creating such a dataset, however, introduces its own set of challenges.
The first is the daunting task of acquiring clinical notes, which is almost impossible for external developers and even challenging for internal developers associated with a health organization due to privacy regulations.
Secondly, even when clinical notes are procured, creating a clinical instruction set necessitates either direct annotation from medical professionals or leveraging external models that have a strong understanding of clinical practices, such as the GPT series \citep{lievin2022can,nori2023capabilities,dash2023evaluation,javaid2023chatgpt}. 
The former approach is laborious and costly, making it impractical for large-scale use, while the latter approach presents the previously mentioned challenges related to privacy and security that are inherent to API-based models.

To address these multifaceted challenges in the clinical settings, we introduce \textbf{Asclepius}, a clinical LLM constructed based on a comprehensive collection of synthetic clinical notes and corresponding instruction-answer pairs. 
These synthetic notes are generated from PMC-Patients \citep{zhao2023pmcpatients}, containing anonymized case reports extracted from PubMed Central, a publicly available biomedical literature archive.
Usage of synthetic notes, unlike real ones, not only enables us to leverage advanced online LLMs to produce comprehensive and high-quality clinical instruction datasets, but also allow for the sharing of these resources and the models trained on them as open-source.
Throughout the entire process of generating these data, we utilized GPT-3.5-turbo, and medical professionals were involved in prompt tuning to ensure the output's clinical accuracy and relevancy.
As a culmination of these efforts, we developed \textbf{Asclepius-7B} and \textbf{Asclepius-13B}, our advanced clinical LLMs capable of handling diverse clinical NLP tasks (see Figure \ref{fig:task}).

We evaluated Asclepius using a rigorous framework that aligns with its intended real-world applications, utilizing \textit{real clinical notes} as our primary evaluation dataset.
For this evaluation, we gathered clinical notes from a diverse set of sources including MIMIC-III \citep{johnson2016mimic}, MIMIC-IV \citep{johnson2023mimic}, i2b2 \citep{uzuner2007evaluating}, CASI \citep{moon2014sense}, and MTSamples\footnote{\url{https://mtsamples.com}}, thereby ensuring a broad coverage of notes from various institutions.
The first goal of our evaluation involved a comparison between our model and GPT-3.5-turbo. 
This comparison allowed us to assess Asclepius's capability to perform at par with API-based LLM across different clinical NLP tasks. 
Additionally, we compared Asclepius against a diverse array of open-source LLMs, including both general domain and clinical-biomedical domain models.
This comparison aimed to validate our model's performance against other locally available LLMs.
We were particularly interested in Asclepius-R, a variant trained with real clinical notes.
Comparing Asclepius with Asclepius-R enabled us to assess the relative performance of models trained with synthetic notes against those trained with real ones.
If a significant performance gap was found, it could possibly challenge our approach's validity.
Hence, this comparison emphasizes the effectiveness of our method in training a clinical LLM using synthetic notes.
In the overall evaluation process, we utilized GPT-4 as an evaluator, which is known to have advanced medical knowledge, to assess the models' performance. 
Furthermore, for the crucial comparison between Asclepius and Asclepius-R, four clinicians were involved in the evaluation to substantiate our claim.

\begin{figure*}[!htbp]
\centering
\includegraphics[width=1.0\textwidth,trim={10 170 30 150}, clip]{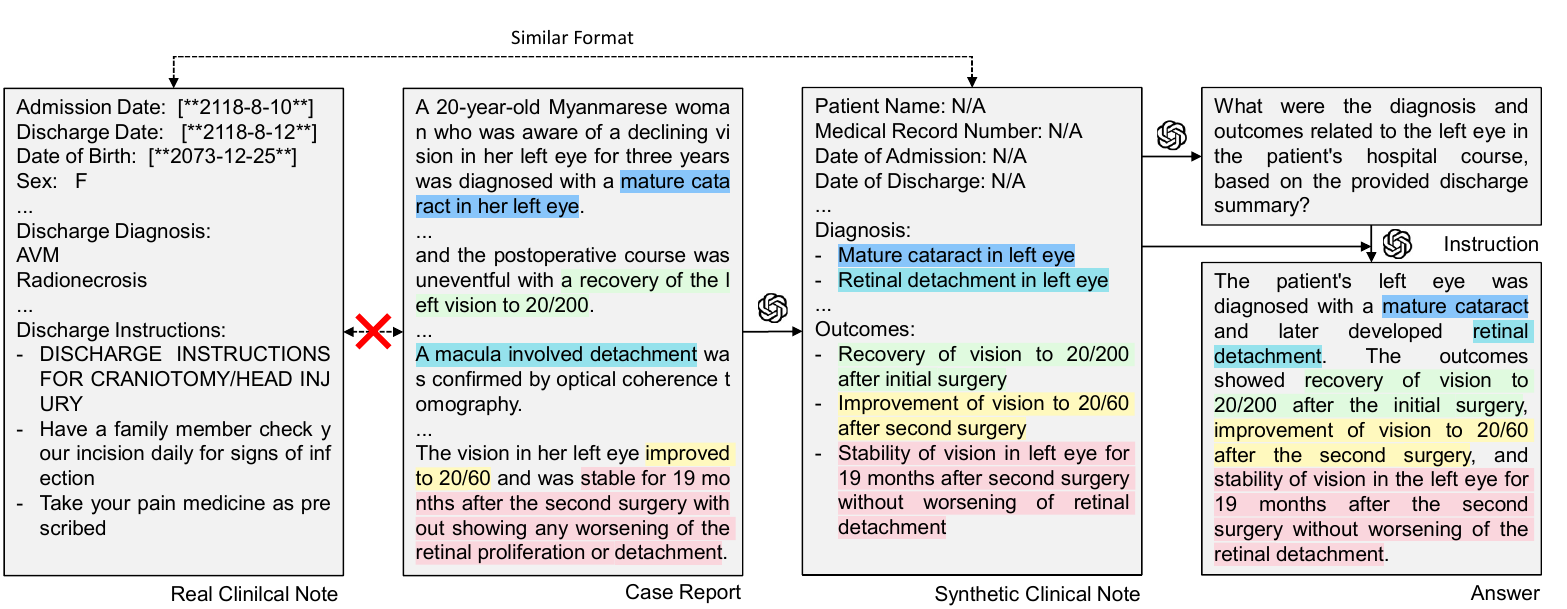}
\caption{The first column is a part of the real discharge summary from MIMIC-III \citep{johnson2016mimic}. Second is a case report from PMC-Patients \citep{zhao2023pmcpatients}, and the third is the synthetic discharge summary created from this case report. Initially, the case report did not resemble the real clinical note in terms of format, but after the transformation, it more closely resembles the real clinical note. At the last column, there is an instruction and answer pair generated from synthetic clinical note. GPT-3.5-turbo was used in all generation processes.}
\label{fig:example}

\vspace{-1em}

\end{figure*}

Our key contributions can be summarized as follows:
\begin{itemize}[leftmargin=4.5mm]

    \item With medical professionals, we created a dataset of 158k synthetic notes each with clinical instruction-answer pairs and have made it publicly available for research.
    

    \item We present our Asclepius-7B and 13B models, trained on synthetic notes. The 13B model shows performance similar to GPT-3.5-turbo on clinical benchmarks.
    

    \item Evaluations using GPT-4 and by medical experts suggest that synthetic clinical notes can be a viable substitute for real ones in building a clinical LLM.

\end{itemize}

By leveraging our methodology, any entity – from healthcare organizations to individual researchers – can develop an LLM capable of understanding and interacting with clinical notes. 
This breakthrough will serve as a crucial stepping-stone to accelerate the research and development of healthcare AI, which has been previously deterred by stringent (yet essential) privacy regulations.

\section{Data Generation}


In Section \ref{2.1}, we discuss the differentiation between clinical notes and case reports and detail how to convert case reports into synthetic clinical notes. 
Section \ref{2.2} delves into extracting specific instruction-answer pairs from these notes for training the clinical LLM. 
Figure \ref{fig:example} illustrates this process with an accompanying example.
It is important to note our method solely uses public data, allowing unrestricted use of LLM (GPT-3.5-turbo). All prompts utilized are listed in Appendix \ref{Appendix A}.

In this research, we specifically focus on the discharge summary, a specific type of clinical note that is extensively used in a variety of clinical tasks.
Henceforth in this paper, the term \emph{clinical note} will specifically refer to the \emph{discharge summary}.

\subsection{Synthetic Clinical Notes} \label{2.1}

Clinical notes are comprehensive records created by healthcare providers to document the care administered to a patient during their stay in a medical facility.
These notes contain sensitive personal health information of patients, and as such, their access and usage are strictly regulated. 
Although public datasets like MIMIC-III \citep{johnson2016mimic} and MIMIC-IV \citep{johnson2023mimic} exist, access is only limited to credentialed individuals, such as those who have completed CITI training. 
This limitation also applies to any products derived from these datasets, such as synthetic data or generative models trained using MIMIC, making it challenging to share them publicly.
On the other hand, case reports are detailed reports on individual patients prepared for academic or educational purposes. 
They are fully anonymized and publicly available through medical journals. 
The contents of a case report mirror that of clinical notes, encompassing admission details, laboratory test results, official diagnoses, and treatment plans. 
Given these similarities, we hypothesized that creating a clinical large language model using case reports would yield a model with performance comparable to one built using authentic clinical notes.
Additionally, this approach would make the model widely accessible without any restrictions, as it would be based on publicly available, anonymized data.

However, using case reports to directly create a large clinical language model as a substitute for real clinical notes presents a problem due to the differences in terms of the characteristics.
Firstly, case reports are written with the intention of being published in academia, thus, they use well-organized and standardized language, whereas clinical notes often contain frequent abbreviations, non-standard terminology, and occasional grammatical errors \citep{lehman2023we}. 
Second, case reports are presented in a continuous narrative form, written in plain text paragraphs. 
Clinical notes, in contrast, are designed for quick referencing by healthcare professionals.
These notes are typically semi-structured through headers such as 'History', 'Physical Examination', 'Assessment', and 'Plan'.
To bridge this gap, we used GPT-3.5 to transform case reports into synthetic clinical notes, giving an instruction to mimic the traits found in real clinical notes. 
Another consideration during this process was the hallucination risk of GPT-3.5 \citep{ji2023survey}. 
Even if the synthetic clinical note closely resembles a real clinical note, any hallucination leading to clinical inconsistency would undermine its validity as a clinical note. 
Therefore, we explicitly specified in the prompt that clinical entities should not be generated in the synthetic clinical notes if they were not mentioned in the case report. 
During the prompt tuning process, clinicians participated and reviewed 50 random samples for each prompt to ensure that the outputs resembled real clinical notes and did not contain inaccuracies or inconsistencies with the original case report.

Consequently, we have obtained 158k high-quality synthetic clinical notes using case reports from PMC-patients dataset \citep{zhao2023pmcpatients}. 
An example of a case report and its converted synthetic clinical note can be found in the Appendix \ref{synthetic_example}.
We used perplexity as a measurement to quantitatively evaluate the similarity of these synthetic clinical notes to real ones.
For this comparison, we further finetuned a pre-trained language model, LLaMA \citep{touvron2023llama}, on a corpus of 57k real discharge summaries from the MIMIC-III database \citep{johnson2016mimic}. 
Then, we measured the perplexity of 200 discharge summaries from three different actual hospital datasets: MIMIC-III (unseen during training), MIMIC-IV \citep{johnson2023mimic}, and i2b2 \citep{uzuner2007evaluating}. 
The MIMIC-III and MIMIC-IV datasets originate from Beth Israel Deaconess Medical Center, whereas i2b2 comes from a different institution, Partners Healthcare. 
We also calculated the perplexity of the 200 case reports from PMC-Patients using the same model.
Finally, we evaluated the perplexity of the synthetic notes transformed from the specific 200 case reports that we had previously measured for perplexity.
Our findings, summarized below, show that the perplexity of real hospital data ranges from 2.186 (in-domain data from MIMIC-III) to 5.178 (data from another hospital, i2b2).
The PMC-Patients' case reports initially had a perplexity of 71.719, but upon transformation into synthetic notes, while preserving the same contents, it dropped to 4.816, thus falling within the range observed for real hospital data.
These results suggest that our synthetic notes likely exhibit a high degree of validity, comparable to real hospital data.

\begin{table}[h]
    \centering
    \setlength{\heavyrulewidth}{1.5pt}
    \resizebox{1\columnwidth}{!}{
        \newcommand{\mytable}{\begin{tabular}{c|c|c|c|c}
\textbf{MIMIC-III} & \textbf{MIMIC-IV} & \textbf{i2b2} & \textbf{PMC-Patients} & \textbf{Synthetic} \\ \hline
2.186              & 2.809             & 5.178         & 71.719                & 4.816        
\end{tabular}}
            \begin{tabular}{c|c|c|c|c}
\textbf{MIMIC-III} & \textbf{MIMIC-IV} & \textbf{i2b2} & \textbf{PMC-Patients} & \textbf{Synthetic} \\ \hline
2.186              & 2.809             & 5.178         & 71.719                & 4.816        
\end{tabular} 
        }
    \vspace{-1em}
\end{table}

\subsection{Clinical Instruction Generation} \label{2.2}

To develop a clinical large language model capable of performing various clinical NLP tasks, a specific training dataset, in the form of instruction-answer pairs, is necessary. 
Considering that our model is targeted towards healthcare professionals, we aimed to incorporate their diverse needs into the instruction sets. 
We initiated the process by defining clinical NLP tasks, based on a comprehensive survey by \citet{wu2022survey}, which analyzed widely used clinical NLP tasks. 
This task list was further refined through consultations with professionals, leading to eight specific task types: Named Entity Recognition, Relation Extraction, Temporal Information Extraction, Coreference Resolution, Question Answering, Abbreviation Expansion, Summarization, and Paraphrasing. 
We created instruction-answer pairs for these eight clinical NLP tasks using GPT-3.5-turbo, based on synthetic clinical notes. 
The method for creating these pairs is as follows.

\begin{enumerate} [leftmargin=4.5mm]
    \item The process started with hand-crafting five professional-verified examples per task as seed data.
    \item These examples, with task type and synthetic clinical note, were input into GPT-3.5-turbo for task-related instruction generation. A bootstrapping method was employed to diversify content by augmenting seed examples with model-generated instructions \citep{wang2022self}.
  \item The generated instructions were fed back into the model along with the notes, prompting the model to generate the corresponding answers. While many studies attempt to generate instructions and answers simultaneously for efficiency \citep{wang2022self, alpaca}, our empirical findings indicated that a sequential generation results in more detailed instructions and answers.
  \vspace{-0.4em}
\end{enumerate}

Employing this approach, we were able to generate high-quality clinical instruction-answer pairs for each synthetic note, culminating in a total of 158,114 pairs. 
Similar to the synthetic notes generation process, physicians were directly involved in the prompt tuning process, thereby ensuring the quality of the instruction-answer pairs. 
Examples of the generated instructions can be found in Appendix \ref{inst}.

\section{Clinical Large Language Model} \label{4}

\subsection{Training}
Recent research \citep{alpaca,vicuna2023,geng2023koala,han2023medalpaca,yunxiang2023chatdoctor,toma2023clinical} has demonstrated the effectiveness of fine-tuning with instruction datasets on foundation language models, such as LLaMA \citep{touvron2023llama}. 
Inspired by these findings, we designed a language model specifically for clinical notes, using LLaMA as the base and incorporating instructions from synthetic clinical notes.
Distinct from other studies, we added an additional step to our process to address a persistent challenge: language models, trained on general domain texts, often struggle to accurately capture the peculiarities found in clinical texts \citep{laparra2020rethinking}.
Previous research has attempted to solve this problem by pre-training base models on clinical notes \citep{alsentzer2019publicly, lewis2020pretrained}. 
Adopting this approach, we applied domain adaptation to LLaMA by pre-training it on synthetic clinical notes before fine-tuning it with clinical instructions. 
Detailed information about the pre-training and instruction fine-tuning processes can be found in Appendix \ref{Experimental Details}.
As a result, we developed two models, \textbf{Asclepius-7B} and \textbf{Asclepius-13B}. 
To our knowledge, these are the first publicly accessible clinical LLMs capable of managing multiple tasks without necessitating task-specific fine-tuning.

\begin{figure*}[t]
\centering
\includegraphics[width=1.0\textwidth]{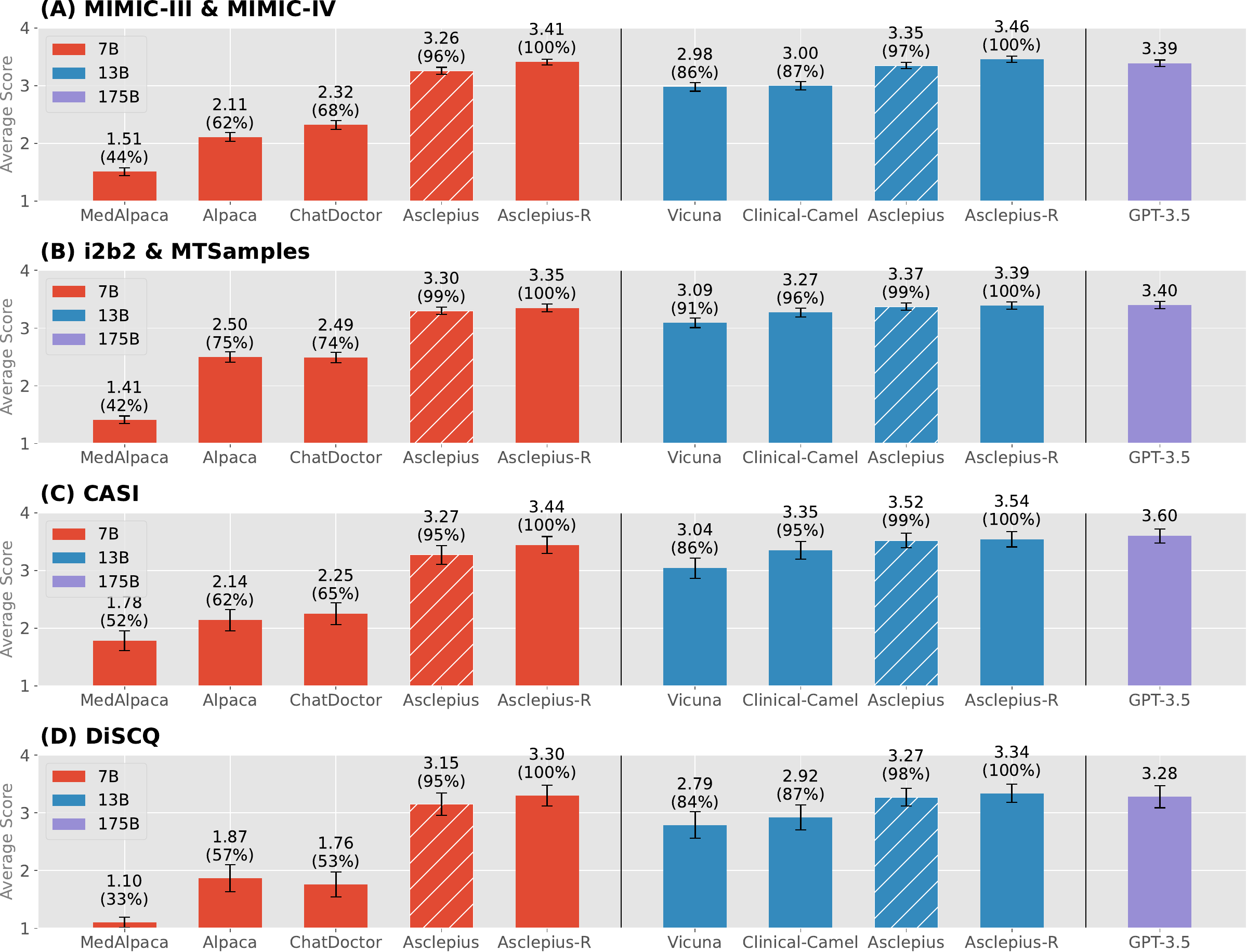}
\caption{The evaluation score from GPT-4 across diverse tasks and models. These tasks include: (A) MIMIC-III and MIMIC-IV (B) i2b2 and MTSamples (C) CASI (D) DiSCQ. The percentages listed beneath the GPT-4 scores represent the ratio of each model's score compared to the highest score achieved within that same model size category. The error bars represent a 95\% confidence interval.}\label{fig:graph}
\vspace{-1em}
\end{figure*}

\subsection{Evaluation} \label{Evaluation}

In our study, we utilized the capabilities of GPT-4 to assess the performance of our models.
GPT-4 has been applied in numerous research as a means to evaluate the results of Natural Language Generation (NLG) models \citep{liu2023gpteval,vicuna2023}.
According to these studies, evaluations derived from GPT-4 – using indicators such as helpfulness and fluency – closely align with human judgment.

However, in the context of the clinical domain, the consequences of mistakes are dire, and there is less tolerance for inaccuracies than in the general domain.
Consequently, we tailored our evaluation criteria to prioritize accuracy, relevancy, and completeness, as any misinformation or omission could potentially result in adverse patient outcomes.
We designed evaluation prompts for GPT-4 to address these specific clinical concerns.
Our clinician-certified four-point scale for scoring is:

\begin{enumerate}[leftmargin=4.5mm]
  \item Unacceptable (1 point): The model's response includes any incorrect or irrelevant contents. If the instruction was unanswerable, the model did not acknowledge this and outputs the wrong answer.
  \item Poor (2 points): The model's response does not contain any incorrect or irrelevant contents but omits significant or crucial contents that the instruction is required for.
  \item Satisfactory (3 points): The model's response does not contain any incorrect or irrelevant contents but omits minor or insignificant contents that the instruction is required for.
  \item Excellent (4 points): The model's response contains all the necessary information that the instruction is requiring for. If the instruction was unanswerable, the model correctly acknowledged this and says that it was unanswerable.
\end{enumerate}
The full prompt can be found in Appendix \ref{Eval_prompt}.

\section{Comparative Analysis}  \label{5}

Despite the various benefits of a clinical LLM trained on \textit{synthetic notes}, a model's ultimate value lies in its performance on \textit{real clinical notes}. 
Accordingly, our evaluation framework employs real discharge summaries as an evaluation dataset, establishing a more authentic and applicable testing ground for our model, Asclepius.
For this evaluation, we gathered clinical notes from a diverse set of sources including MIMIC-III \citep{johnson2016mimic}, MIMIC-IV \citep{johnson2023mimic}, i2b2 \citep{uzuner2007evaluating}, MTSamples, and CASI \citep{moon2014sense}, thereby ensuring a broad coverage of notes from various institutions.

We conduct a comparative study to analyze the performance of Asclepius against several others using GPT-4 evaluation specified in Section \ref{Evaluation}. 
Our initial point of comparison is GPT-3.5-turbo, wherein we aim to ascertain whether Asclepius can match the performance and versatility of API-based LLM in various clinical NLP tasks.
We also include an evaluation of other open-source instruction fine-tuned LLMs that are trained on general domain data, such as Alpaca \citep{alpaca} and Vicuna \citep{vicuna2023}, and those tailored for the clinical-biomedical domain, such as MedAlpaca \citep{han2023medalpaca}, ChatDoctor \citep{yunxiang2023chatdoctor}, and Clinical-Camel \citep{toma2023clinical}. 
Including these models aims to compare Asclepius's performance on clinical NLP tasks with other locally available models, thereby validating our methodology in developing our model.

Lastly, we include Asclepius-R, a variant of our model, trained using 57k real clinical notes from the MIMIC-III dataset \citep{johnson2016mimic}\footnote{To use MIMIC-III in conjunction with online API-based language models like GPT, one must comply with specific guidelines (https://physionet.org/news/post/415). We strictly followed these guidelines when conducting experiments involving real clinical notes.}.
Asclepius-R, having been both pre-trained and fine-tuned on these real notes, is directly compared to Asclepius, our model trained on 158k synthetic notes.
This comparison allows us to explore the performance of models developed with synthetic data in relation to those trained with real data.
It is important to clarify that our objective is not to claim that synthetic notes can completely replace real ones, but rather to show that a model trained on synthetic notes can be a viable alternative to one trained on real data.
To optimize their performances, we leveraged the maximum amount of data available for each model.
The performances of Asclepius and Asclepius-R, when trained on datasets of the same size, are detailed in the ablation study in Appendix \ref{ablation study}.
By including Asclepius-R in our analysis, we can thoroughly assess the potential of our method in training large language models using synthetic clinical notes instead of real ones.

\subsection{Preliminary Evaluation}\label{Preliminary Assessment}

We conducted a comparative analysis of our model, Asclepius, with Asclepius-R, GPT-3.5-turbo, and other open-source instruction-tuned large language models.
The initial performance assessment involved MIMIC-III (unseen during training Asclepius-R) and MIMIC-IV discharge summaries, which are in-domain data for training Asclepius-R which comes from the same health institution. 
We then extracted instruction data from these summaries to compile a test set, following the methodology outlined in Section \ref{2.2}.
As shown in Figure \ref{fig:graph}-(A), Asclepius, trained on synthetic notes, demonstrated performance closely aligned with that of Asclepius-R, which was trained on in-domain data.
Moreover, when we conducted the same evaluation on i2b2 notes and MTSamples (Figure \ref{fig:graph}-(B)), which originated from different institutions and were of types not used in the training of Asclepius-R, the performance gap between Asclepius and Asclepius-R narrowed for both 7B and 13B models further.

Another key observation is that Asclepius outperforms all open-source LLMs and even exhibits performance comparable to GPT-3.5-turbo.
However, it is important to consider that the test set, created from the aforementioned discharge summaries, followed the same process used for the training sets of Asclepius and Asclepius-R. 
This could potentially bias the comparison in their favor.
To ensure a fairer comparison, we broadened our evaluation to directly employ prompts that were used in \citet{agrawal2022large}, which addresses Coreference Resolution and Abbreviation Expansion tasks on the CASI dataset \citep{moon2014sense}.
As illustrated in Figure \ref{fig:graph}-(C), even for previously unseen types of prompts, the Asclepius model 1) outperformed all other open-source LLMs and 2) displayed performance closely aligned with GPT-3.5 for the 13B model.
This pattern is consistent across all individual benchmarks, detailed in Appendix \ref{Comprehensive Experimental Results}.

\subsection{Practical Evaluation}\label{DiSCQ}
Designed for use by professionals in actual healthcare settings, it is crucial to test clinical LLM's effectiveness on actual queries posed by healthcare professionals. 
As such, we utilized the DiSCQ dataset \citep{lehman2022learning} - a set of clinician-posed questions derived from MIMIC-III discharge summaries - for practical evaluation.
However, since the authors of the DiSCQ dataset allowed clinicians to annotate questions freely while reading the discharge summaries, without providing specific guidance, it is often impossible to find answers to the questions within the corresponding discharge summaries. 
This presents a significant challenge when evaluating a model's performance in answering these questions.
To mitigate this issue, we first used GPT-4 to filter the dataset, tasking it with identifying any evidence within the discharge summary that could potentially answer a given question. 
We then randomly selected 100 questions from this filtered dataset for our evaluation.
Refer to Appendix \ref{apd:discq} for more detail.

The results depicted in Figure \ref{fig:graph}-(D) confirm that the performance on questions annotated by real clinicians shows the same pattern as before. 
Our model, Asclepius, demonstrated significant superiority over other baseline models. 
In the case of the 13B model, its performance was on par with GPT-3.5-turbo, despite being ten times smaller in model size.
Moreover, when compared with Asclepius-R, the performance remains comparable.
Based on these findings, it may be suggested that building a clinical LLM from real patient notes - which poses a privacy risk - might not be necessary.
It is possible that a model with similar performance could be achieved using synthetic notes.

\begin{figure}[t]
\includegraphics[width=0.5\textwidth, trim={190 100 150 100}, clip]{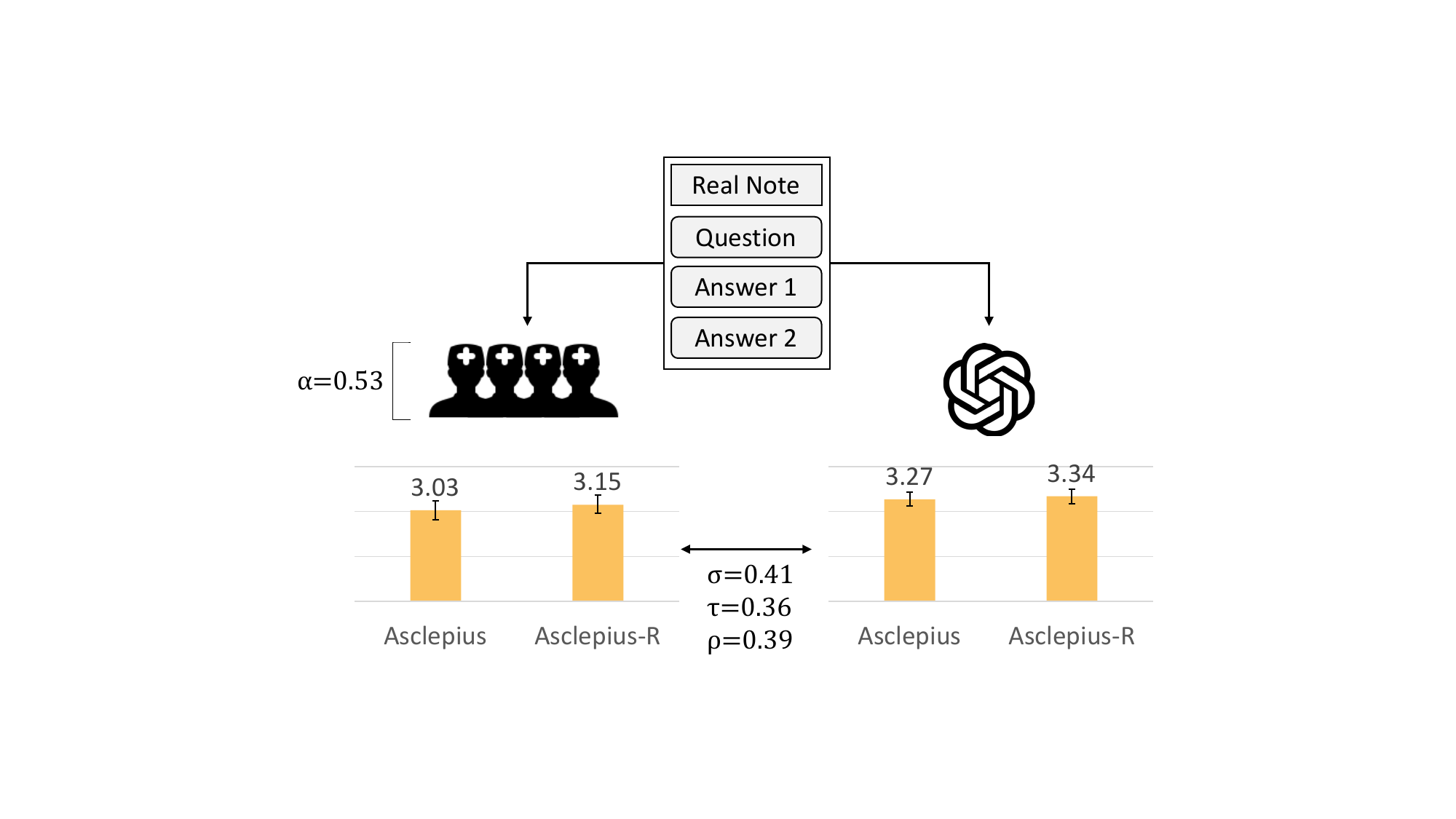}
\caption{Professional and GPT-4 evaluation of Asclepius-13B and Asclepius-R-13B responses to 100 DiSCQ questions, featuring inter-professional Krippendorff's alpha ($\alpha$) agreement and GPT-4 to professional average alignment via Pearson, Kendall-Tau, and Spearman coefficients ($\sigma,\tau, \rho$).
The error bars represent a 95\% confidence interval.
}
\label{fig:eval}
\vspace{-1em}
\end{figure}

\subsection{Professional Evaluation}\label{Professional Assessment}

Despite GPT-4's advanced medical knowledge, boasting an accuracy level of 86\% on the United States Medical Licensing Examination \citep{nori2023capabilities}, it is uncertain whether the conclusions drawn by GPT-4 match with those of actual healthcare professionals. 
Considering that professionals are the most likely users of our model, it's necessary to validate our previous conclusion that a model trained on synthetic notes performs comparably to one trained on real notes, involving these professionals.

To address this, we solicited evaluations from healthcare professionals for Asclepius-13B and Asclepius-R-13B, on DiSCQ dataset.
Concurrently, we measured the alignment of professionals' evaluations to that of GPT-4, thus bolstering the validity of our previous evaluations.
The evaluation was carried out by a team of four clinicians. 
We asked them to rate the quality of responses generated by the two models (with criteria in \ref{Evaluation}) to the same 100 questions from the DiSCQ dataset that were used in Section \ref{DiSCQ}.
We ensured that each question was evaluated by at least two experts, allowing us to also assess inter-rater agreement among them. 
The overall process and its result are visualized in Figure \ref{fig:eval} and the user interface employed for this process can be seen in Appendix \ref{ui}.

Our statistical analysis revealed a Krippendorff's alpha ($\alpha$) of 0.53. 
As a measure of agreement among evaluators, this value signifies a moderate level of inter-annotator agreement \citep{landis1977measurement}, offering preliminary assurance of the credibility of our evaluations.
The clinicians assigned average scores of 3.03 and 3.15 to Asclepius-13B and Asclepius-R-13B, respectively.
We conducted a paired sample t-test on the evaluations from the clinicians, comparing Asclepius and Asclepius-R. 
The result did not reject the null hypothesis stating that the performance of the two models is equivalent (p-value = 0.18).
Additionally, when a statistical test of the same kind was applied to the scores provided by GPT-4, the null hypothesis could not be rejected in this case either (p-value = 0.40). 
While the interpretation of these results is limited by the sample size of 100, it nonetheless offers a preliminary conclusion that the performance of the two models is approximately similar.
When comparing the alignment of GPT-4 and the professionals' evaluations, we found a moderate level of Pearson ($\sigma=0.41$), Kendall-Tau ($\tau=0.36$), and Spearman ($\rho=0.39$) correlation coefficients \citep{landis1977measurement}, lending further validity to our previous experiments that were solely evaluated by GPT-4.

\section{Related Work}\label{Related}

\subsection{Synthetic Clinical Notes}


Efforts to create synthetic clinical notes include \citet{melamud2019towards} using LSTM \citep{hochreiter1997long} on MIMIC-III discharge summaries, \citet{ive2020generation} employing transformer architecture \citep{vaswani2017attention} with MHR \citep{perera2016cohort} and MIMIC-III databases, \citet{li2021synthetic} using text generation models like GPT2 \citep{radford2019language} on the i2b2 2010 \citep{uzuner20112010} and n2c2 2018 datasets \citep{henry20202018} for data augmentation, and \citet{zhou2022datasiftertext}'s BERT-based method \citep{devlin2018bert} for de-identifying MIMIC-III clinical records.
All of these synthetic notes were derived from real hospital data, which implies certain constraints on their usage.
Distinctively, our approach harnesses publicly accessible case reports for generating synthetic notes. 
Thus, our synthetic data does not possess the limitations seen in the aforementioned works. 
Consequently, models trained on our data are free from such constraints, making them shareable with the public.


\subsection{Language Models for Clinical NLP tasks}

There have been several clinical language models developed, each designed to address a specific clinical NLP tasks.
Notable examples of such models include ClinicalBERT \citep{alsentzer2019publicly}, Clinical-Longformer \citep{li2023comparative}, Gatortron \citep{yang2022large} which based on transformer encoder structure, and Clinical-T5 \citep{lehman2023we} which is based on transformer encoder-decoder structure. 
All these models are pre-trained using clinical notes and then fine-tuned for each specific task.  
While this approach has shown to be effective, the limited size of these models restricts their ability to perform multiple tasks simultaneously. 
This limitation reduces their practicality in real-world scenarios, as it is more convenient to address various tasks with a single model rather than managing multiple models specialized for each task.
Asclepius is the first clinical large language model that is capable of handling multiple clinical NLP tasks.

\section{Conclusion}\label{Conclusion}

In this paper, we present \textbf{Asclepius}, trained on 158k high-quality synthetic clinical notes and instruction sets.
Evaluations across diverse benchmark datasets against other LLMs demonstrate that Asclepius performs on par with GPT-3.5-turbo while locally executable in hospital settings.
Most importantly, it exhibits no significant disparity with models trained on actual clinical notes, thereby validating the use of synthetic notes for training clinical large language models. 
The evaluations were not solely reliant on GPT-4 but also involved appraisals by four clinicians, reinforcing the validity of our conclusions. 
For future research, all synthetic data, model weights, and code used in these experiments are publicly available.
This opens the door for not only healthcare institutions but also businesses and researchers to develop clinical large language models. 
We believe this has the potential to significantly advance healthcare AI, especially in areas previously held back by privacy concerns.

\section*{Limitations}
This study has several limitations that should be acknowledged.
Firstly, our model was primarily designed and tested only on discharge summaries, which may limit its application and generalizability to other types of clinical notes, such as progress notes, nursing notes, or radiology notes. 
Future research should aim to develop a model that can effectively function with a wider variety of note types. 
Secondly, our model is currently only capable of handling one-turn instruction following tasks. 
This may constrain its use in more dynamic and interactive healthcare settings where conversations between the model and healthcare professionals are required for a comprehensive understanding of the patient's clinical notes. 
We plan to extend this model in future studies to allow interactive dialogues, thereby increasing its utility and applicability in real-world clinical scenarios.
Third, we initially used GPT for data generation, but its terms of use prohibit using its output to train models for business competition.
However, with the recent advances in open-source LLMs, this issue could be addressed by replacing GPT's role with one of them.
Lastly, but most importantly, we did not extensively investigate the model's hallucination capacity, which may affect its reliability and accuracy when implemented in practice. 
Our model can generate hallucinated responses, which may cause critical issues in practical applications (see Appendix \ref{apd:hallucination}).
It is important to note that the current model is intended for research purposes and should not yet be used in actual clinical practice. 
Further research is required to rigorously test and enhance the model's performance and ensure its safe and effective use in clinical settings.

\section*{Acknowledgements}
This work was supported by Institute of Information \& Communications Technology Planning \& Evaluation (IITP) grant (No.RS-2019-II190075), National Research Foundation of Korea (NRF) grant (NRF-2020H1D3A2A03100945, RS-2023-00262527), funded by the Korea government (MSIT), and NAVER Digital Bio Innovation Research Fund, funded by NAVER Corporation (Grant No.3720230020).

\bibliography{custom}
\clearpage
\appendix

\section{Prompts} \label{Appendix A}
All experiments employing GPTs were executed using the HIPAA-compliant GPT model available on Azure\footnote{\url{https://learn.microsoft.com/en-us/azure/compliance/offerings/offering-hipaa-us}}.
The creation of synthetic notes, as well as the generation of instruction-answer pairs based on these notes, amounted to a total cost of \$1000. 
Generating instructions and answers involving MIMIC-III discharge summaries required an additional \$500.
The evaluation process, which was conducted using GPT-4, incurred a total cost of \$200.
In the following section, we provide all the prompts employed in our GPT experiments.

\subsection{Synthetic Note Generation}\label{synthetic note generation}
\begin{tcolorbox}
\small
You are an intelligent clinical model.\\

[The start of case report]

\textcolor{red}{\{Case Report\}}

[The end of case report]\\

Based on the patient's case report provided, please generate a synthetic discharge summary in the style of an Electronic Health Record (EHR).\\

Please follow these requirements:\\

1. Generate only the discharge summary. Do not generate any other phrases such as notification.\\
2. If there are any standard clinical terms used in the case report, they should be replaced with their commonly used non-standardized equivalents in the discharge summary. For example, "hypercholesterolemia" can be rewritten as "high cholesterol".\\
3. The discharge summary includes abbreviations that are not defined in the context.\\
4. The discharge summary may contain minor grammatical errors.\\
5. Ensure that the discharge summary does not contain any clinical information or details (such as medication names, dosages, treatment plans, diagnoses, procedures, test results, etc.) that are not explicitly mentioned or defined within the given case report.\\
6. While preserving the structure of the EHR and maintaining medical consistency, generate a detailed and comprehensive discharge summary.\\
7. The discharge summary is a comprehensive document that can be organized using several distinct headings.\\
8. For patients who have not yet been discharged, create a section in the discharge summary summarizing their medical progression.
\end{tcolorbox}

\vfill\eject
\subsection{Clinical Instruction Generation} 
\begin{tcolorbox}
\small
You are a healthcare professional who often uses an advanced clinical language model that can analyze clinical notes. You ask the questions and use the outputs from this model to aid in your decision-making process.\\

Let's suppose you want to ask a question about the task of "\textcolor{red}{\{Task\}}" related to the given discharge summary. How would you frame your question?\\

[Discharge Summary Begin]

\textcolor{red}{\{Discharge Summary\}}

[Discharge Summary End]\\

We also offer five example questions related to the task for your reference. Although you can refer to these samples, try to create a distinct query. The crucial aspect of your question is that it should be answerable using only the information available in the discharge summary. Also, try to keep your question as brief and straightforward as possible.\\

[Example Questions Begin]

\textcolor{red}{\{Examples\}}

[Example Questions End]\\

Please output solely a question.
\end{tcolorbox}

\subsubsection{Examples - Named Entity Recognition}
\begin{tcolorbox}
\small
"Can you identify and categorize all the medical conditions"\\
"Could you extract all medications prescribed and their corresponding dosages?"\\
"What were the diagnostic tests and their results?"\\
"Identify all the medical procedures"\\
"Can you recognize and list any lifestyle factors such as smoking, alcohol consumption, or exercise habits"
\end{tcolorbox}

\subsubsection{Examples - Relation Extraction}
\begin{tcolorbox}
\small
"Identify the connection between the medication 'Metformin' and the patient's diagnosed condition 'Type 2 diabetes'"\\
"Establish the relation between the prescribed dosage of 'Lisinopril' and the patient's 'hypertension' management"\\
"Can you find the link between the use of 'antibiotics' and the patient's 'post-operative infection'"\\
"Determine the association between the patient's lifestyle modifications, specifically 'diet, and exercise', and the improvement in 'cholesterol levels'."\\
"Decipher the relationship between the 'radiation therapy' administered and the progress of the patient's 'breast cancer'"
\end{tcolorbox}

\vfill\eject

\subsubsection{Examples - Temporal Information Extraction}
\begin{tcolorbox}
\small
"Can you extract the duration of the patient's treatment"\\
"Identify the duration of the hospital stay mentioned in the discharge summary."\\
"Retrieve the timestamps of any surgical procedures mentioned in the discharge summary."\\
"Extract the date and time of the last medication administration recorded in the discharge summary."\\
"Identify any temporal references to follow-up appointments or scheduled tests in the discharge summary."
\end{tcolorbox}

\subsubsection{Examples - Coreference Resolution}
\begin{tcolorbox}
\small
"Which medication does 'it' refer to in the line mentioning 'it should be taken twice daily' in the medication instructions section of the discharge summary?"\\
"Please clarify what 'this procedure' refers to in the surgeon's notes section of the discharge summary."\\
"Identify the coreferents for the pronouns used in the second paragraph of the discharge summary."\\
"In the patient education section of the discharge summary, when 'these exercises' are mentioned, what specific exercises are being referred to?"\\
"Who does 'he' refer to in the sentence 'He is expected to recover fully' in the prognosis section of the discharge summary?"
\end{tcolorbox}

\subsubsection{Examples - Question Answering}
\begin{tcolorbox}
\small
"Considering the 'uncontrolled diabetes' statement in the hospital release notes, what lifestyle changes and medication revisions can be recommended?"\\
"Identify all the instances suggesting 'adverse reactions' from drugs mentioned in the discharge synopsis."\\
"In light of the 'congestive heart failure' diagnosis in the patient's discharge summary, what are the subsequent tests and procedures that need to be arranged?"\\
"Locate all references to 'dietary restrictions' in the discharge document and provide an explanation for each constraint."\\
"Based on the 'chronic kidney disease' mentioned in the discharge documents, what routine follow-up strategy and patient awareness should be put into action?"
\end{tcolorbox}

\subsubsection{Examples - Abbreviation Expansion}
\begin{tcolorbox}
\small
"What is the expanded form of the abbreviation 'COPD'"\\
"Could you decode all the abbreviations present in this discharge report?"\\
""HTN" has been mentioned frequently in this report. Could you clarify what it means?"\\
"In the context of this medical report, what would "CAD" typically stand for?"\\
"I see 'DM' used here in relation to a patient's condition, could you provide the full term?"
\end{tcolorbox}

\vfill\eject

\subsubsection{Examples - Summarization}
\begin{tcolorbox}
\small
"Can you provide a succinct summary of the key clinical findings and treatment recommendations outlined in this discharge summary?"\\
"Can you identify and condense any lifestyle and medication modifications recommended in the patient's discharge summary?"\\
"Given the patient's discharge summary, can you extract the diagnosis and prognosis information and summarize it in layman's terms for the patient's understanding?"\\
"Could you extract and summarize the patient's progress during hospitalization, as well as key notes regarding her discharge planning?"\\
"What were the key findings from the lab tests, imaging, and other diagnostic procedures? Please summarize these in simple terms."
\end{tcolorbox}

\subsubsection{Examples - Paraphrasing}
\begin{tcolorbox}
\small
"The discharge summary states that the patient suffered from 'an anomalous blockage in the coronary artery.' Could you paraphrase this medical term into simpler language that the patient might understand?"\\
"How would you rephrase the line in the discharge summary, 'Patient exhibits signs of acute rhinosinusitis,' to make it easier for a non-medical professional to grasp?"\\
"In this discharge summary, it mentions 'diabetes mellitus type 2 with hyperglycemia.' Can you provide a paraphrase that might be more straightforward for the patient and their family?"\\
"The term 'post-operative seroma' appears in the patient's discharge summary. Can you paraphrase this to a less clinical terminology?"\\
"Could you translate the sentence, 'The patient's condition was complicated by acute renal failure due to ischemia,' into more common terms to aid in communicating the situation to the patient?"
\end{tcolorbox}

\vfill\eject

\subsection{Answer Generation}
\begin{tcolorbox}
\small
You are an intelligent clinical language model.\\

[Discharge Summary Begin]

\textcolor{red}{\{Discharge Summary\}}

[Discharge Summary End]\\

[Instruction Begin]

\textcolor{red}{\{Instruction\}}

[Instruction End]\\

Above, we provide you with a part of the discharge summary and the instruction that the healthcare professional gave about it.
Generate a response to the healthcare professional's instruction using the given discharge summary.

Here are the requirements:\\
- Your response must be accurate and concise to the instruction.\\
- If the instruction is not fully answerable within the given discharge summary, explain the reason why it is unanswerable using the given information.\\
- Do not say that you cannot respond as an AI model.\\
- Do not ask back nor rephrase the instruction.\\

Response:
\end{tcolorbox}

\vfill\eject

\subsection{Evaluation} \label{Eval_prompt}
\begin{tcolorbox}
\small
You are an intelligent clinical language model. \\

[Discharge Summary Begin]

\textcolor{red}{\{Discharge Summary\}}

[Discharge Summary End]\\

[Instruction Begin]

\textcolor{red}{\{Instruction\}}

[Instruction End]\\

[Agent A's Answer Begin]

\textcolor{red}{\{A\}}

[Agent A's Answer End]\\

[Agent B's Answer Begin]

\textcolor{red}{\{B\}}

[Agent B's Answer End]\\
\vdots\\

Above, we provide you with a discharge summary and the instruction that the healthcare professional gave about the discharge summary.
You are also provided with \textcolor{red}{\{Number of Samples\}} corresponding responses from \textcolor{red}{\{Number of Samples\}} different clinical models.
Your task is to read the discharge summary and the instruction carefully, then find the answer to the instruction.
Then, compare your answer with each model's response and evaluate the response based on the following criteria.

Criteria : 

1. Unacceptable (1 point): The model's response includes any incorrect or irrelevant content. If the instruction was unanswerable, the model did not acknowledge this and outputs the wrong answer.

2. Poor (2 points): The model's response does not contain any incorrect or irrelevant content but omits significant or crucial content that the instruction is required for.

3. Satisfactory (3 points): The model's response does not contain any incorrect or irrelevant contents but omits minor or insignificant contents that the instruction is required for.

4. Excellent (4 points): The model's response contains all the necessary information that the instruction is requiring for. If the instruction was unanswerable, the model correctly acknowledged this and says that it was unanswerable.

When evaluating each score based on the above criteria, ensure that each judgment is not affected by the other model's response.

The first line must contain only \textcolor{red}{\{Number of Samples\}} values, which indicate the score for each model, respectively.
The \textcolor{red}{\{Number of Samples\}} scores are separated by a space.
Output scores without explanation.
\end{tcolorbox}



\vfill\eject

\onecolumn

\section{Synthetic Note Example} \label{synthetic_example}
\subsection{PMC-Patients}
\begin{figure*}[!htbp]
    \centering
    \begin{tcolorbox}
    \small
    \input{synthetic_note/1_before}
    \end{tcolorbox}
\end{figure*}

\newpage

\subsection{Synthetic Note} 
\begin{figure*}[!htbp]
    \begin{tcolorbox}
    \small
    \input{synthetic_note/1_after}
    \end{tcolorbox}
\end{figure*}

\vfill\eject

\section{Sample Generated Instructions}\label{inst}

\begin{table*}[!htp]\centering
\caption{Sample of Generated Instructions using GPT-3.5-turbo}
\scriptsize
\resizebox{\textwidth}{!}{
\begin{tabular}{lrr}\toprule
\textbf{Task} &\textbf{Instruction} \\\cmidrule{1-2}
Named Entity Recognition &What medical procedures were performed on the patient during her hospital course, as mentioned in the discharge summary? \\\cmidrule{1-2}
Abbreviation Expansion &What abbreviation was expanded using the acronym "ANH" in the diagnosis section of the discharge summary? \\\cmidrule{1-2}
Relation Extraction &What radiologic findings support the diagnosis of inner ear hemorrhage (ILH) in the given discharge summary? \\\cmidrule{1-2}
Temporal Information Extraction &When was the patient started on oral acyclovir and what was the duration of treatment? \\\cmidrule{1-2}
Coreference Resolution &What is the referent for 'the lesion' that underwent complete surgical excision, in the treatment section of the discharge summary? \\\cmidrule{1-2}
Paraphrasing &Can you simplify the language used in the discharge summary to make it more understandable for the patient and their family? \\\cmidrule{1-2}
Summarization &Can you summarize the patient's hospital course, treatment, and diagnoses according to the given discharge summary? \\\cmidrule{1-2}
Question Answering &What was the reason for the patient's transfer to ICU and what was the treatment plan for infection-induced respiratory failure? \\\midrule
\bottomrule
\end{tabular}
}
\end{table*}


\section{Individual Experimental Results} \label{Comprehensive Experimental Results}


\begin{table*}[!htp]\centering
\caption{Individual Experimental Results. Bolded entries denote the best scores, while underscored ones represent the second-best results for each model size.}\label{tab: res}
\scriptsize
\resizebox{\textwidth}{!}{
\setlength{\tabcolsep}{0.3em}
\begin{tabular}{lrrrrrrrrrrrr}\toprule
\multicolumn{2}{c}{Model Size} &\multicolumn{5}{c}{7B} &\multicolumn{4}{c}{13B} &175B \\\cmidrule(lr){1-2} \cmidrule(lr){3-7} \cmidrule(lr){8-11} \cmidrule(lr){12-12} 
Dataset &Sample Size &MedAlpaca &Alpaca &ChatDoctor &Asclepius &\makecell[t]{Asclepius\\-R} &Vicuna &\makecell[t]{Clinical\\-Camel} &Asclepius &\makecell[t]{Asclepius\\-R} &\makecell[t]{GPT-3.5\\-turbo} \\\cmidrule{1-12}
MIMIC-III &413 &1.45 &1.95 &2.21 &\ul{3.21} &\textbf{3.43} &2.90 &2.93 &\ul{3.36} &\textbf{3.48} &3.46 \\\cmidrule{1-12}
MIMIC-IV &500 &1.57 &2.24 &2.42 &\ul{3.31} &\textbf{3.40} &3.05 &3.06 &\ul{3.34} &\textbf{3.45} &3.34 \\\cmidrule{1-12}
i2b2 &619 &1.38 &2.47 &2.43 &\ul{3.25} &\textbf{3.32} &3.08 &3.25 &\ul{3.34} &\textbf{3.35} &3.36 \\\cmidrule{1-12}
MTSamples &101 &1.61 &2.73 &2.85 &\ul{3.57} &\textbf{3.58} &3.21 &3.39 &\ul{3.51} &\textbf{3.61} &3.64 \\\cmidrule{1-12}
CASI (AE) &100 &1.39 &1.86 &1.94 &\ul{2.97} &\textbf{3.17} &2.89 &3.15 &\textbf{3.36} &\textbf{3.36} &3.72 \\\cmidrule{1-12}
CASI (CR) &100 &2.16 &2.41 &2.56 &\ul{3.56} &\textbf{3.70} &3.18 &3.54 &\ul{3.67} &\textbf{3.71} &3.47 \\\cmidrule{1-12}
DisCQ &100 &1.10 &1.87 &1.76 &\ul{3.15} &\textbf{3.30} &2.79 &2.92 &\ul{3.27} &\textbf{3.34} &3.28 \\\midrule
\bottomrule
\end{tabular}
}
\end{table*}

\newpage

\twocolumn

\section{Experimental Details}\label{Experimental Details}
While numerous prior studies, using LLaMA \cite{touvron2023llama} as their basis, have restricted sequence length to 512 due to resource constraints \citep{alpaca, han2023medalpaca, yunxiang2023chatdoctor}, we recognized the average token length of MIMIC-III discharge summaries \citep{johnson2016mimic} to be around 1400. 
Thus, we adjusted the maximum sequence length in our model to 2048. 
To counterbalance the increased memory usage due to the extended context length, we incorporated efficient training methodologies \citep{dao2022flashattention,chen2016training}, as utilized in Vicuna \citep{vicuna2023}. 
Our model underwent one pre-training epoch using notes, followed by three epochs of instruction fine-tuning. 
In both stages, the following hyperparameters were used: a learning rate of 2e-5 and a global batch size of 128. 
For the configuration of additional hyperparameters, we adhered to the standards used in Alpaca \citep{alpaca}. 
All experiments were conducted using either 8x A100 80G GPUs or 8x A6000 48G GPUs.

\section{Ablation Study}\label{ablation study}
\begin{figure}[!htbp]
    \centering
    \includegraphics[width=0.47\textwidth]{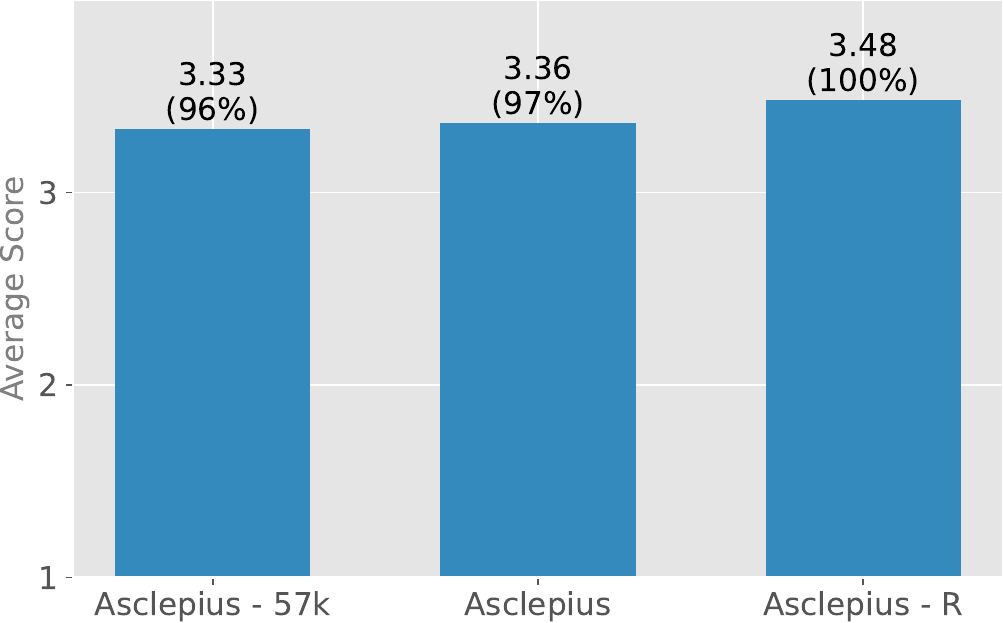}
    \caption{Ablation Study on MIMIC-III (test set) Instructions}
    \label{fig:ablation}
\end{figure}

To gauge the effect of data size on model performance, we generated Asclepius-57k, a model built using an equal number of synthetic notes as the MIMIC-III notes used in Asclepius-R. 
We set out to compare the performance of Asclepius-57k with Asclepius and Asclepius-R at the 13B model scale. 
Upon analyzing the results, as outlined in Figure \ref{fig:ablation}, we observed a slight decrease in performance for Asclepius-57k compared to Asclepius.

\section{DiSCQ} \label{apd:discq}

The DiSCQ \citep{lehman2022learning} dataset is composed of 1k questions annotated by physicians as they encountered triggering terms while reading MIMIC-III discharge summaries \citep{johnson2016mimic}.
An example of a triggering term along with the triggered question can be found below Table \ref{tab:discq}.
It's important to note that the dataset only provides the physicians' questions, and the corresponding answers are left unknown.
To utilize the DiSCQ dataset for our model evaluation, we followed three steps:

\begin{enumerate}
    \item We reformatted the original questions into "\textit{\{Question\} with respect to \{Trigger\}}", as the authors of DiSCQ did.
    \item We presented discharge summaries and reformatted questions to GPT-4, asking it to determine whether there were supporting evidence of the questions present in the summaries. This resulted in a filtered set of 200 questions.
    \item Out of these 200 questions, we randomly sampled 100 for a practical and professional evaluation.
\end{enumerate}

\begin{table}[h]
    \centering
    \resizebox{0.5\textwidth}{!}{
    \begin{tabular}{>{\centering}p{2cm}|p{8cm}}
    \hline
    Discharge Summary &CV : The patient 's family refused coronary artery catheterization . The patient was given ASA , Plavix , heparin drip x 24 hours , nitro drip , atorvastatin , metoprolol , and lisinopril . Her chest pain was controlled with morphine . Her SBP remained in the 160s-170s on hospital day 1 and she was gently diuresed . On hospital day 2 she experienced \textcolor{red}{atrial fibrillation} with HR in the 140s. Her metoprolol dose was increased from 25 mg PO bid to 50 mg PO bid ... \\ \hline
    Triggering Term &\textcolor{red}{atrial fibrillation}\\ \hline
    Triggered Question &Were interventions done? \\ \hline
    Reformatted Question &Were interventions done with respect to \textcolor{red}{atrial fibrillation}?\\ \hline
    \end{tabular}
    }
    \caption{DiSCQ Example}
    \label{tab:discq}
\end{table}%

\onecolumn
\section{User Interface for Professional Evaluation}\label{ui}
\begin{figure}[H]
    \centering
    \fbox{
    \includegraphics[width=0.75\textwidth,trim={30 180 200 30}, clip]{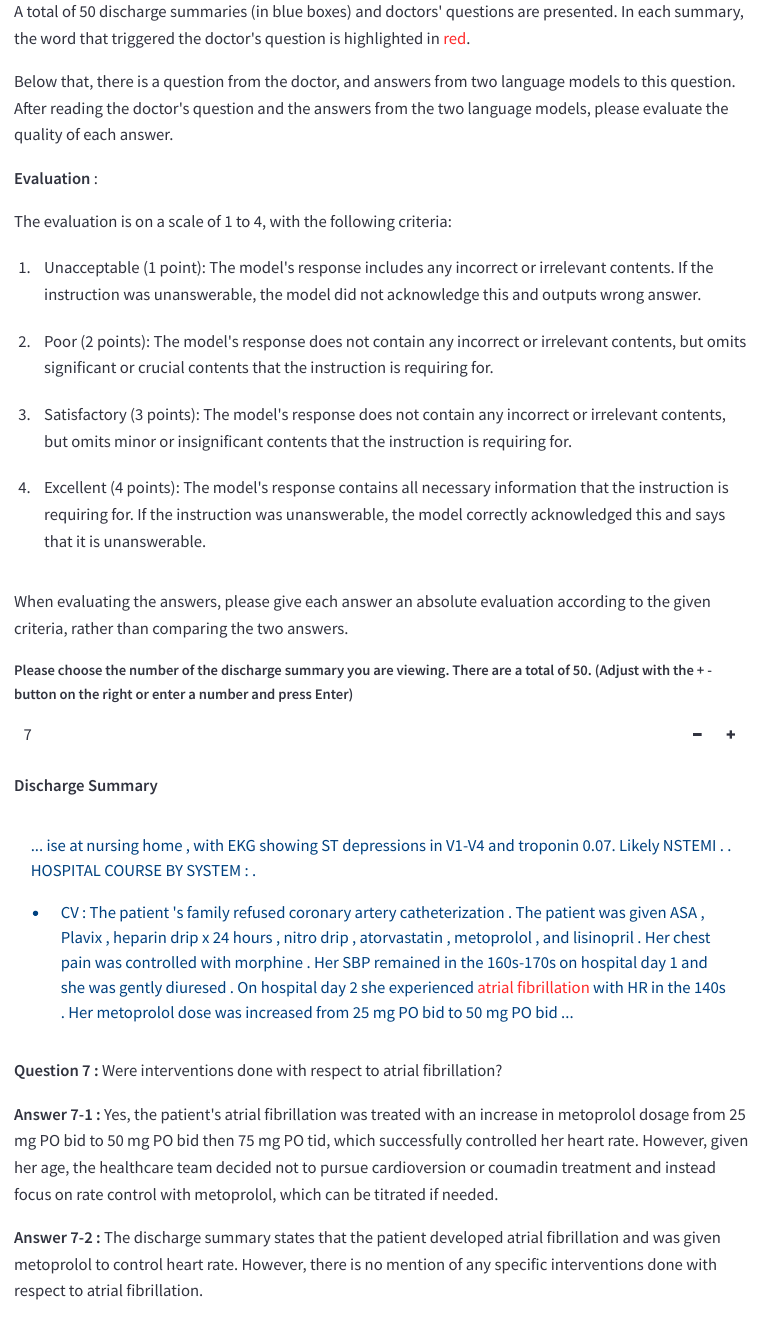}
    }
    \caption{User Interface for Professional Evaluation}
    \label{fig:ui}
\end{figure}

\newpage

\section{Hallucination Analysis}\label{apd:hallucination}
\newlength{\width}
\setlength{\width}{0.11\linewidth}

\begin{table*}[!ht]
    \centering
    \scriptsize
    \caption{Model Hallucination Analysis Results. The phrases that directly relate to hallucination are bolded.}
    \label{tab:hallucination}
    \setlength{\tabcolsep}{0.32em}
    \renewcommand{\arraystretch}{1.2} 
    \begin{tabular}{|>{\raggedright}p{0.9\width}|>{\raggedright}p{1.3\width}|>{\centering}p{0.6\width}|>{\centering}p{0.6\width}|>{\raggedright}p{1.5\width}|>{\raggedright}p{1.3\width}|>{\raggedright\arraybackslash}p{2.2\width}|}
    \hline
        \textbf{Category} & \textbf{Description} & \textbf{No. of Cases \\(Asclepius)} & \textbf{No. of Cases\\(Asclepius\\-R)} & \textbf{Related Note Sample} & \textbf{Question Sample} & \textbf{Response Sample} \\ \hline
        Non-Existent Entity Mention & The model references an entity that does not exist within the document. & 4 & 4 & On scoping him he had significant supraglottic edema. & Did the significant supraglottic edema impact his airways/breathing and were any interventions? & Yes, the significant supraglottic edema impacted the patient's airways/breathing as revealed by the hoarse voice and the need for b.i.d. laryngoscopy. \textbf{However, the edema resolved with time without further intervention.}\\ \hline
        Ambiguous Response & The model provides an answer that is unclear or can be interpreted in multiple ways. & 2 & 3 & - & What is the clinical evidence of elevated intracranial pressure and brainstem compression? & The clinical evidence of elevated intracranial pressure and brainstem compression was noted on the post-intubation examination, where the patient showed signs of increased intracranial pressure and brainstem compression. \\ \hline
        Incorrect Clinical Interpretation & The model offers a clinical interpretation that is incorrect or inappropriate based on the given data. & 4 & 3 & 9. History of syncope and first degree A-V heart block with beta blocker therapy. & What is the indication for beta blocker therapy in the case of first degree A-V heart block? & The indication for beta blocker therapy in the case of first degree A-V heart block is \textbf{to prevent symptomatic bradycardia.} \\ \hline
        False Non-Existence Assertion & The model incorrectly asserts that an existing entity in the document does not exist. & 2 & 0 & In the Deanna, \textbf{Tmax 202.8}, BP 200 - 240/60 s, HR 80 s, RR 20 s, Sat 75\% on RA in triage, up to 200\% on NRB. & Has there been any recorded temperature regarding the new fever? & There is no recorded temperature regarding the new fever in the given discharge summary. \\ \hline
    \end{tabular}

\end{table*}

We additionally conducted an analysis about the hallucination capability of Asclepius-13B and Asclepius-R-13B models.
In our evaluation criteria for the model, a score of 1 indicates that the model's response contains incorrect or irrelevant content. This score indirectly assesses the model's tendency to produce hallucinated responses. We focused our investigation on responses from within the DisCQ dataset that consistently received a score of 1 from all clinicians involved. We categorized them into four groups, as illustrated in the table available at the Table \ref{tab:hallucination}.

In the case of Asclepius-13B, 12 out of 100 responses received a score of 1. Conversely, Asclepius-R-13B, which was trained on real clinical notes, had 10 responses with the same score. 
Notably, 9 of these cases were common between the two models and were similarly categorized. 
This similarity in the pattern of hallucination between the model trained on synthetic clinical notes and the one trained on real clinical notes also strengthens our key argument.

\end{document}